%% file: JINT_R1.tex
\documentclass[pdflatex,sn-mathphys]{sn-jnl}

\usepackage[acronym,nomain,nonumberlist]{glossaries}
\usepackage{pgfplots}
\pgfplotsset{compat=newest}
\usepackage{tikz}
\usetikzlibrary{positioning}
\usetikzlibrary{shapes.geometric}
\usetikzlibrary{shapes.misc}
\usetikzlibrary{external}
\usetikzlibrary{shapes,arrows,patterns}
\usetikzlibrary{arrows.meta}
\usetikzlibrary{plotmarks}
\usetikzlibrary{calc,decorations.markings,math}
\usepackage{bm}
\usepackage[utf8]{inputenc}
\usepackage[english]{babel}
\newtheorem{assumption}{Assumption}
\usepackage[]{units}
\newlength\fwidth

\usepackage{threeparttable}
\usepackage{xcolor}
\hypersetup{
    colorlinks,
    linkcolor={black!50!black},
    citecolor={black!50!black},
    urlcolor={black!80!black}
    }

\usepackage{tikz}
\usetikzlibrary{arrows.meta}
\usetikzlibrary{calc,decorations.markings,math}
\usetikzlibrary{external}
\usetikzlibrary{plotmarks}
\usetikzlibrary{positioning}
\usetikzlibrary{shapes,arrows,patterns}
\usetikzlibrary{shapes.geometric}
\usetikzlibrary{shapes.misc}
\usetikzlibrary{external}

\input{abbreviations}

\jyear{2021}%

\begin{document}

\title[Multi-Stage NMPC under Varying Communication Delays for MAV]{Multi-Stage NMPC for a MAV based Collision Free Navigation under Varying Communication Delays}


\author*[1]{\fnm{Andreas} \sur{Papadimitriou}}\email{andpap@ltu.se}
\equalcont{These authors contributed equally to this work.}

\author[1]{\fnm{Hedyeh} \sur{Jafari}}\email{hedjaf@ltu.se}
\equalcont{These authors contributed equally to this work.}

\author[2]{\fnm{Sina Sharif} \sur{Mansouri}}\email{sina.sharif.mansouri@scania.com}

\author[1]{\fnm{George} \sur{Nikolakopoulos}}\email{geonik@ltu.se}

\affil[1]{\orgdiv{Department of Computer, Electrical and Space Engineering}, \orgname{Robotic \& AI team}, \orgaddress{Lule\r{a} University of Technology, \postcode{SE-97187}, \country{Sweden}}}

\affil[2]{\orgdiv{Autonomous Driving Lab}, \orgname{Scania Group}, \orgaddress{\city{Södertälje}, \postcode{SE-15139}, \country{Sweden}}}




\abstract{Time delays in communication networks are one of the main concerns in deploying robots with computation boards on the edge. 
This article proposes a multi-stage \gls{nmpc} that is capable of handling varying network-induced time delays for establishing a control framework being able to guarantee collision-free \glspl{mav} navigation. 
This study introduces a novel approach that considers different sampling times by a tree of discretization scenarios contrary to the existing typical multi-stage \gls{nmpc} where system uncertainties are modeled by a tree of scenarios.
Additionally, the proposed method considers adaptive weights for the multi-stage \gls{nmpc} scenarios based on the probability of time delays in the communication link. 
As a result of the multi-stage \gls{nmpc}, the obtained optimal control action is valid for multiple sampling times. 
Finally, the overall effectiveness of the proposed novel control framework is demonstrated in various tests and different simulation environments.}

\glsresetall

\keywords{Multi-stage NMPC, MAV, Network delays, Navigation}
\maketitle

%
%
%
\section{Introduction}


The last decade the \glspl{mav} have steadily gained interest in the fields of real-life applications, such as infrastructure inspection~\cite{mansouri2018cooperative}, underground mine tunnel inspection~\cite{mansouri2020deploying}, and bridge inspection~\cite{hallermann2014visual}. 
The key objective in all these use-cases is the online collection of critical information like images, 3D models, and other sensorial data to create safer conditions for the personnel while reducing the overall inspection time.

Fully autonomous performance of the \glspl{mav} is one of the key challenges in deploying them in real-world applications, while it should overcome uncertainties in localization, limited on-board computation power, delays in control layers, dynamic/static obstacles, etc. Meanwhile, advances in technologies, such as 5G~\cite{ullah20195g} telecommunications technology, enable the use of cloud and edge computing for \gls{mav} applications. 
In this case, the heavy computational processing for multiple processes, such as mapping, localization, and path planning can be carried on the edge computing side, while retaining a fast bi-directional link with the MAV. However, one of the main challenges in such networked applications is the limited bandwidth, the time delays, and the overall package losses that degrade the overall control performance and could lead the system to instability.~\cite{NPT08}. 
 
This article proposes a novel \gls{nmpc} framework to address the communication delays in the network. In the proposed method, multi-stage \gls{nmpc} considers the scenario trees with different sampling times to deal with the network's varying delays. The proposed method takes into account the non-linear dynamics of the \gls{mav} while the global map and obstacles position are assumed known. The probability of the time delays is considered as an adaptive weight in the scenarios of the multi-stage \gls{nmpc}. Thus, prioritizing the scenario with a higher probability and resulting in collision-free navigation. 
\subsection{Background \& Motivation}
Many research approaches have considered the use of a multi-stage \gls{nmpc} framework in the process industry~\cite{holtorf2019multistage}, where the presence of stochastic model uncertainties can lead to significant control performance degradation or, in the worst case, directly affect the feasibility of the operation. However, in the robotics community and more specifically in the use-case of aerial robots, the mathematical translation model of the \gls{mav} can comprehensively follow the kinematic behavior of the \glspl{mav}~\cite{alaimo2013comparison, alaimo2013mathematical}.
Therefore, the main focus is surrounding awareness based on the equipped sensor suite. To that end, mapping, localization, navigation, and control are the main components for mission accomplishment. Yet, the risk of one of these modules fail is high, for example, the numerous path planning challenges, such as obstacles, uncertainty in localization, noise, biases, time delays, and other stochastic events, increasing the risk of collision and jeopardize the mission. The proposed framework considers network-induced time delays towards the safe navigation of an aerial platform with embedded obstacle avoidance capabilities.

\begin{figure*} [htbp!] \centering
\includegraphics[width=0.99\linewidth]{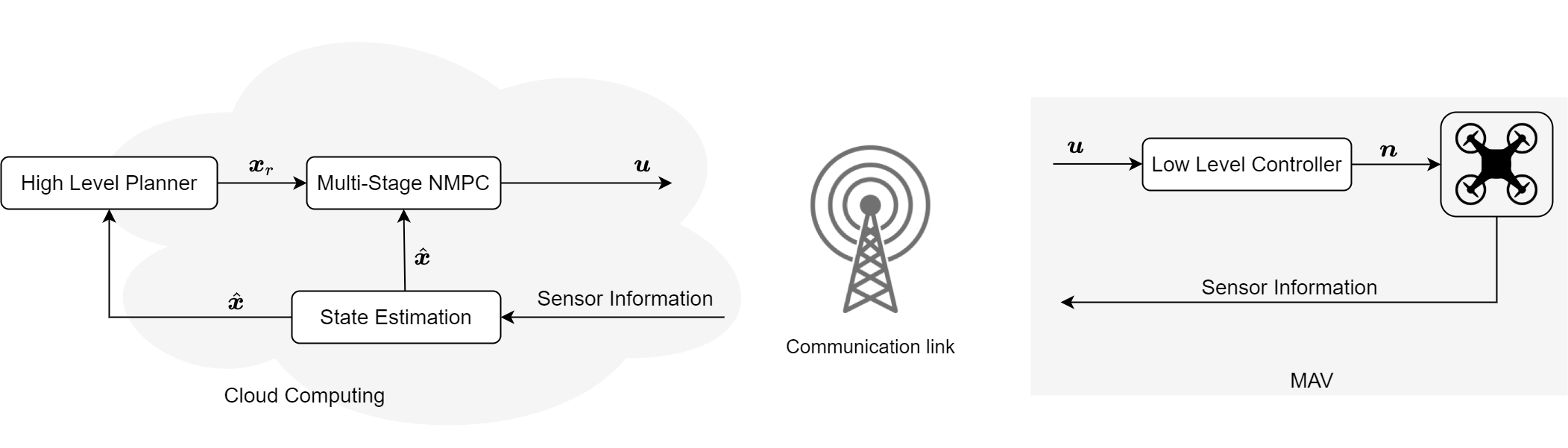}
\caption{Control scheme of the proposed multi-stage \gls{nmpc} module, where the control actions $\bm{u}$ based on sensor information is obtained in the cloud, and the low-level controller on the \gls{mav} generates motor commands $\bm{n}$. The $\bm{x}$ is state of the system, $\bm{x}_r$ is reference value, and $\hat{\bm{x}}$ is estimated state.}
\label{fig:controlartichecture}
\end{figure*}

Towards designing a computationally efficient controller that predicts further in the future, a theoretical framework for establishing an adaptive prediction horizon was investigated in~\cite{KRENER201831}, where an \gls{ahmpc} has been developed with a varying step prediction horizon that depended on the deviation from the operating point. Thus, the further from the desired states the larger horizons steps were considered. Similarly, to decrease the computation burden of the online optimization, in~\cite{sun_multi}, an event-based \gls{mpc} with an adaptive prediction horizon strategy was proposed for the tracking of a unicycle robot. The authors proposed a control scheme that reduces the solving rate when the robot is near the desired location. However, the event-based mechanism relied on the error threshold between the current and reference states that reduced the prediction horizon and the solving frequency and thus could not foresee further in the prediction horizon. A study that fused varying the prediction horizon and time-varying delays was introduced in~\cite{Arai}, where the authors proposed a method for utilizing only resources that are available at a specific time instant by adjusting the prediction horizon properly. The idea of predicting future driving trajectory based on uncertainties for different time horizons with the use of multiple sensors to create a robust and trustworthy prediction system was studied in~\cite{Huang}.


The topic of \gls{mav} navigation and path planning is well studied in the related literature~\cite{lavalle2006planning, galceran2013survey}. 
Exploration algorithms like the frontier exploration algorithms~\cite{fentanes2011new}, entropy-based algorithms~\cite{burgard2005coordinated}, and information-gain algorithms~\cite{bhattacharya2013distributed} provide a global planning strategy for the \gls{mav}, while an additional reactive control layer provides a local obstacle avoidance to prevent collisions with the environment. 
The most widely used reactive control layer is the artificial potential fields~\cite{barraquand1992numerical}, while another approach that has received attention in the last years is the \gls{nmpc}~\cite{kamel2017nonlinear}, which has been applied to perform real-time obstacle avoidance for \glspl{mav}~\cite{mansouri2020nmpcmine} and disturbance rejection~\cite{papadimitriou_nmhe}.
However, all of these methods consider fixed time steps, while in real-life applications and especially in networked enabled \glspl{mav} the feature of a time-varying path planning and a corresponding controller that can take into consideration time variations is vital for collision-free navigation.
\subsection{Contributions}

The first contribution stems from developing a multi-stage \gls{nmpc} for considering a tree of sampling times while providing collision-free paths for the \gls{mav}. The varying sampling time addresses the random delays in the communication link between the \gls{mav} and the edge, which results in time uncertainties in the control actions period. In this article, the varying communication delays in the information exchange between the \gls{mav} and the controller are addressed by the multi-stage \gls{nmpc} that is defined by a scenario tree for different sampling times. To the best of our knowledge, no work so far has considered different sampling times in the multi-stage \gls{nmpc} framework.

The second contribution stems from the introduction of adaptive weights for each scenario of the proposed multi-stage \gls{nmpc}. The adaptive weights are assigned based on the varying uncertainties of time delays as they are stochastic in the communication link. This approach results in realistic control strategies based on real-world limitations.

The final contribution stems from the overall application on a \gls{mav} use-case and the corresponding extensive analysis of the control framework performance under iterative simulations. As it will be presented in the sequel, the proposed control scheme enables collision-free navigation in an obstacle environment, while the generated path distance is reduced in comparison to a fixed sampling time \gls{nmpc} scheme for path planning.

\subsection{Outline}
The rest of this article is structured as follows. In Section~\ref{sec:Problem Statement} the research problem is defined, while highlighting the challenges and limitations. Section~\ref{sec:nmpc} provides the non-linear model of a \gls{mav}, the theoretical control framework and the formulation of the optimization problem. The simulation specifics and results are provided in Section~\ref{sec:result}, while concluding remarks and future work are discussed in~\ref{sec:conclusions}.
\section{Problem Statement} \label{sec:Problem Statement}
When a networked control scheme is considered for the control of a \gls{mav} the network-induced time delays directly affect the state estimation and the control actions sampling time. Thus this article will propose a multi-stage \gls{nmpc} that considers the varying sampling times and establish a collision-free path planner. Fig.~\ref{fig:controlartichecture} illustrates the proposed concept while highlighting the overall system architecture. The high-level planner generates references $\bm{x}_r$ for the multi-stage \gls{nmpc}, the controller provides actions $\bm{u}$ for the low-level controller, and the low-level controller feeds the motor commands $\bm{n} = \{\bm{n}_i , i\in \mathbb{N}^{\ge 4}\}$ to the \gls{mav}.
%
Without losing generality, the following assumptions are made:

\begin{assumption}
Delays caused by communication are  bounded in a specific known range and can be mitigated by different sampling times, as they affect the control action period.
\end{assumption}
\begin{assumption}
The delays affecting the system are transmission delays in the communication link and not processing delays.
\end{assumption}
\begin{assumption}
The transmission delays between the low-level controller and the system are negligible. 
\end{assumption}

\begin{assumption}
An NMPC control action $\bm{u}$ that satisfies the constraints for sampling times $t_{s}^{i}$ and $t_{s}^{i+1}$ with $i\in\mathbb{N}$ will satisfy the constraints for all sampling times within $[t_{s}^{i},t_{s}^{i+1}]$.
\end{assumption}

\section{Multi-Stage Nonlinear Model Predictive Control} \label{sec:nmpc}
\subsection{MAV Dynamics} \label{sec:mavdynamic}
In this article, the six \gls{dof} dynamics of a \gls{mav} are considered as defined in the body frame (Fig.~\ref{fig:coordinateframes}) and modelled by \eqref{eq:modeluav}:
\begin{figure} [htbp!] \centering
\includegraphics[width=0.95\linewidth]{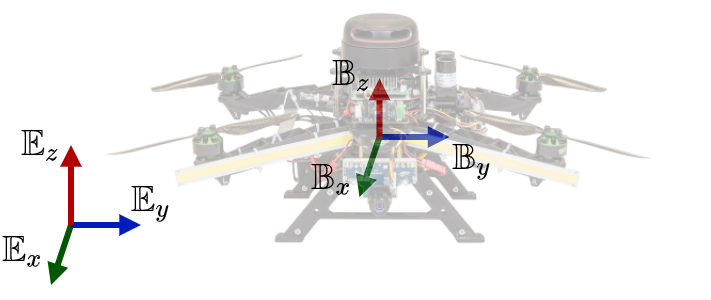}
\caption{Illustration of the \gls{mav} with the attached body fixed frame $\mathbb{B}$ and inertial frame $\mathbb{E}$.}
\label{fig:coordinateframes}
\end{figure}
\begin{subequations} \label{eq:modeluav}
\begin{align}
        \dot{\bm{p}}(t) &= \bm{v}(t) \\ 
        \dot{\bm{v}}(t) &= \bm{R}_{x,y}(\theta,\phi) 
        \begin{bmatrix} 0 \\ 0 \\ T \end{bmatrix} + 
        \begin{bmatrix} 0 \\ 0 \\ -g \end{bmatrix} - 
        \begin{bmatrix} A_x & 0 & 0 \\ 0 &  A_y & 0 \\ 0 & 0 & A_z \end{bmatrix} \bm{v}(t),   \\
        \dot{\phi}(t) & = \nicefrac{1}{\tau_\phi} (K_\phi\phi_d(t)-\phi(t)),  \\
        \dot{\theta}(t) & = \nicefrac{1}{\tau_\theta} (K_\theta\theta_d(t)-\theta(t)),
\end{align}
\end{subequations}
where $\bm{p}=[p_x,p_y,p_z]^\top \in \mathbb{R}^3$ is the position vector and $\bm{v} = [v_x, v_y, v_z]^\top \in \mathbb{R}^3$ is the vector of linear velocities, 
$\phi,\theta \in \mathbb{R} \cap [-\pi,\pi]$ are the roll and pitch angles respectively, $\bm{R}_{x,y}$ is the rotation matrix about 
the $x$ and $y$ axes, $T \in [0,1] \cap \mathbb{R}$ is the mass-normalized thrust, $g$ is the gravitational acceleration, 
$A_x, A_y$ and $A_z \in \mathbb{R}$ are the normalized mass drag coefficients.
The low-level control system is approximated by first-order dynamics driven by the reference roll and pitch angles $\phi_d$ and $\theta_d$ with gains of $K_\phi, K_\theta \in \mathbb{R}^+$ and time constants of $\tau_\phi, \tau_\theta \in \mathbb{R}^+$ respectively.

\subsection{Objective Function}
The system states are $\bm{x}=[\bm{p},\bm{v}, \phi,\theta]^\top$ and the control input is denoted by $\bm{u}=[T,\phi_d,\theta_d]^\top$. To obtain a  discrete-time dynamical system at time instance $k \in \mathbb{Z}^+$, the model expressed by \eqref{eq:modeluav} is discretization by the Euler method and with a sampling time of $t_s$ as:
\begin{equation}
\bm{x}_{k+1} = f(\bm{x}_k, \bm{u}_k).
\end{equation}

The \gls{nmpc} approach solves a finite-horizon problem at every time instant $k$ with the prediction horizon of $N \in \mathbb{N}^{\ge 2}$. The states and control actions are expressed by $\bm{x}_{k+j+1|k}$, and $\bm{u}_{k+j\mid k}$ respectively for $k+j,\,\forall j \in \{0,1, \dots, N-1\}$ steps ahead from the current time step $k$. The purpose of the \gls{nmpc} is the tracking of a reference state $\bm{x}_r=[\bm{p}_r,\bm{v}_r,\phi_r,\theta_r]^\top$ by generating the desired thrust $T$ and angles $\phi_d$, $\theta_d$ commands for the attitude controller, while guaranteeing a safety distance from a priori known obstacles.
For this purpose, the finite horizon cost function can be written as:
%
%
\begin{multline} \label{eq:costfunction} 
J = \sum_{j=0}^{N-1} 
  \underbrace{\|\bm{x}_{k+j+1|k}-\bm{x}_{r}\|_{\bm{Q}_x}^2}_\text{tracking error} \\     
   + \underbrace{\|\bm{u}_{k+j+1|k}-\bm{u}_{r}\|_{\bm{Q}_u}^2}_\text{actuation} 
   + \underbrace{\|\bm{u}_{k+j|k}-\bm{u}_{k+j-1|k}\|_{\bm{Q}_{\Delta u}}^2 }_\text{smoothness cost}.
\end{multline} 
The objective function consists of three terms. The first term ensures the tracking of the desired states $\bm{x}_{r}$ by minimizing the deviation from the current states. The second term, penalizes the deviation from the hover thrust with horizontal roll and pitch, where $\bm{u}_{r}$ is $[g,\, 0,\, 0]^\top$.
The last term tracks the aggressiveness of the obtained control actions. In addition, the weights of the objective function's terms are denoted as $\bm{Q}_x \in \mathbb{R}^{8\times 8}$, $\bm{Q}_u\in \mathbb{R}^{3\times 3}$ and $\bm{Q}_{\Delta u}\in \mathbb{R}^{3\times 3}$ respectively, which reflect the relative importance of each term.
\subsection{Constraints}
\subsubsection{Cylinder Obstacles}
There are different types of obstacles in the surrounding environment, nonetheless all these types of obstacles can be categorized to three types as cylindrical shapes, polytope surfaces, and constrained entrances~\cite{lindqvist2020non}. In this work, we mainly target the cylinder-shaped obstacles. The constraints are defined in parametric form and their positions are fed directly to the \gls{nmpc} scheme \cite{ESmall}. When the \gls{mav} is outside the obstacle, the associated cost is forced to be zero and for that purpose the function $max(h,0)= [h]_+$ is utilized. With the proper selection of the $h$ expression, the constrained area is negative outside the obstacle and positive inside of it.

In case of a cylinder obstacle, the two equations for the safety radius $h_c$ and the maximum altitude allowance $h_{z_{max}}$ are defined based on the center position $(x_{obs},y_{obs})$, the radius $r_{obs}$ and the height $z_{obs}$ of the cylinder as follows: 
\begin{subequations} 
\begin{align}
& h_{c} = r^{2}_{obs} - (x_{k + j|k} - x_{obs})^2 - (y_{k+j|k}-y_{obs})^2, \label{eq:cylinderconstrainta} \\ 
& h_{z_{max}} = z_{obs} - z_{k+j|k}.  \label{eq:cylinderconstraintb}
\end{align}
\end{subequations}
By multiplication of \eqref{eq:cylinderconstrainta} and \eqref{eq:cylinderconstraintb} the cylinder constraint is defined as:
\begin{equation}\label{eq:cylinderconstraints}
    C_{c} = \sum_{j=0}^{N}[h_c]_+[h_zmax]_+ = 0.
\end{equation}

\subsubsection{Input Constraint}

Additionally the inputs are constrained within specific boundaries in the following form:
\begin{equation} \label{eq:input_constraints}
    \bm{u}_{min} \leq \bm{u} \leq \bm{u}_{max},
\end{equation}
where the lower bound $\bm{u}_{min} = [T_{min}, \phi_{min}, \theta_{min}]^\top$ and the upper bound $\bm{u}_{max} = [T_{max}, \phi_{max}, \theta_{max}]^\top$ denote the minimum and maximum values the control actions can take. 




\subsection{Multi-Stage NMPC}

\begin{figure} [htbp!] \centering
\includegraphics[width=0.9\linewidth]{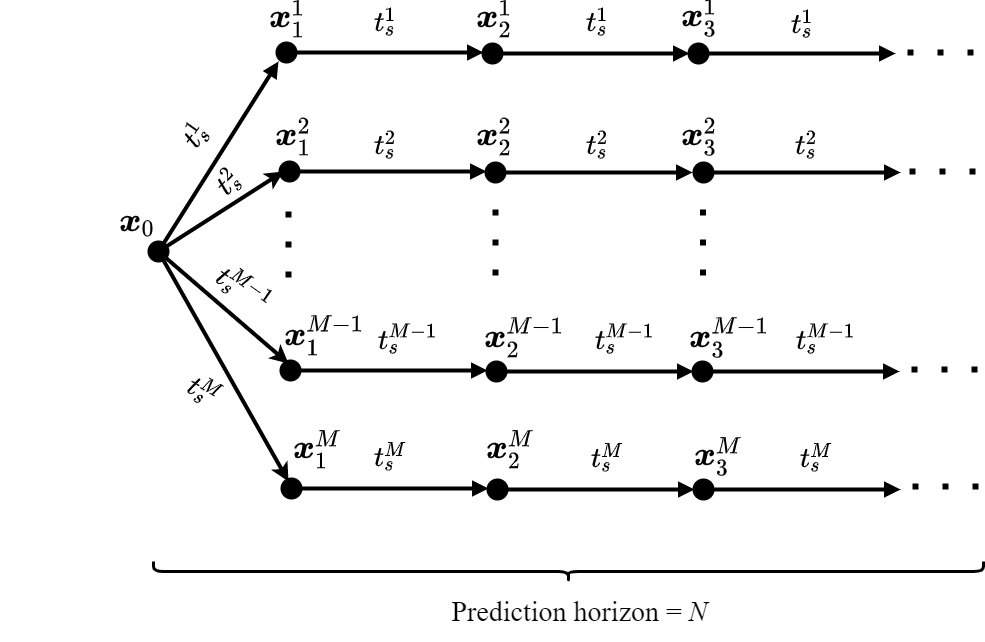}
\caption{Illustration of the scenario tree with different sampling times.}
\label{fig:nmpcstructure}
\end{figure}

Most of the multi-stage \gls{nmpc} approaches model the uncertainties by a tree of  scenarios~\cite{lucia2012new}. In our case, each scenario is defined based on different sampling times $t_s^i, i \in \{1, \dots, M\}$. To that end, we assume that the sampling time is bounded in range of $[t_s^{min}, t_s^{max}]$, which  means that the upper and lower bounds are defined. Based on these bounds and our defined sampling times, each branch of the tree is made as depicted in Fig.~\ref{fig:nmpcstructure}. Our method varies drastically from the traditional move blocking strategy~\cite{Morari_move_blocking} for model-based control problems that fixes the input for several time steps. In the proposed multi-stage framework, the $\bm{u}$ is the same in all branches of the tree, which means that the obtained control action can satisfy all the sampling times.  

The multi-stage \gls{nmpc} objective function based on the classic \gls{nmpc} cost~\eqref{eq:costfunction} denoted here as $J_i$ for the $M \in \mathbb{N}$ scenarios can be formulated as follows:
\begin{subequations}\label{eq:mainoptProb}
\begin{alignat}{2}
&\!\min_{\{u_{k+j\mid k}\}^{N-1}_{j=0}}       & &\sum_{i=1}^{M}  \omega_i J_i \quad  \forall i \in \{1, \dots, M\}\label{eq:optProb}\\
&\text{s.t.} &      & \bm{x}_{k+j+1\mid k}=f_{i}(\bm{x}_{k+j\mid k},\bm{u}_{k+j\mid k}),\label{eq:constraint1}\\
&                  &      & \text{Constraints}~ \eqref{eq:cylinderconstraints},  \eqref{eq:input_constraints}. \label{eq:constraint5}
\end{alignat}
\end{subequations}
%
To consider uncertainties in delays, the term $\omega_i$ defines the weight of each scenario objective function of the multi-stage \gls{nmpc}. The $\omega_i$ is updated based on the probability of the delays. To estimate the $\omega_i$ adaptive term, the network transmission delays are stored in a finite buffer as follows:
\begin{multline}
   \bm{t_d}=\{(t_{d,1}, \dots, t_{d,l}), l \in \{k-(n_{max}-1),\dots,k\}\} ,
\end{multline}
where $t_d$ is a single communication delay and $n_{max}$ is the limited window size of stored network delays.
Furthermore, it is assumed that the communication delays that affect the control framework are following a \textit{Gamma} distribution. Initially, the shape $\alpha$ and scale parameters $\beta$ of the \textit{Gamma} distribution are calculated from the mean $\mu$ and the standard deviation $\sigma$ of set $\bm{t_d}$ as:
\begin{equation}
\alpha = \frac{\mu^2}{\sigma^2}, ~~ \beta = \frac{\sigma^2}{\mu}.
\end{equation}
The probability of a single random delay $t^i_d$ from a \textit{Gamma} distribution with parameters $\alpha,\beta$ falls in the interval $[0,t_{d}]$ and is given from the \gls{cdf} \cite{mathematicshandbook} as:
\begin{equation}
    G(t_{d};\alpha,\beta) = \frac{1}{\beta^{\alpha}\Gamma(\alpha)}\int_{0}^{t_{d}} x^{a-1}e^{\frac{-x}{\beta}}dx,
\end{equation}
%
where $\Gamma(\cdot)$ is the Gamma function  \cite{mathematicshandbook}. Based on the obtained probabilities for the selected span of delays, the $\omega_i$ for each branch of the tree is derived as follows:
%
%
%
\begin{equation} \label{eq:qualityfunction_new}
\omega_i = 
\begin{cases}
G(t_d^1;\alpha,\beta)      &  \,  \text{if } i = 1\\
G(t_d^i;\alpha,\beta) - G(t_d^{i-1};\alpha,\beta),  &  \,  \text{if } i \in  \{2, \dots, M\}
\end{cases}
\end{equation}
%
%
%
%

At each time step, the multi-stage \gls{nmpc} generates an optimal sequence of control actions $\bm{u}_{k|k}^{\star}$, $\dots$, $\bm{u}_{k+N-1|k}^{\star}$, and only the first control action $\bm{u}_{k|k}^\star$ is applied based on a zero-order hold element as $\bm{u}(t)=\bm{u}_{k|k}^\star$ for $t\in [kt_s, (k+1)t_s]$. 

The developed multi-stage \gls{nmpc} with 3D collision avoidance constraints, is solved with \gls{panoc}~\cite{sathya2018embedded} to guarantee a real-time performance. In the related literature for solving the multi-stage \gls{nmpc} optimization problem, algorithmic procedures are used \cite{Lucia2013} relying on solving each of the scenarios independently until the solutions converge and the constraints are satisfied. To this extent, a large optimization problem can be solved in a reasonable amount of time. In contrast, in this article, we attempt a one-shot solution by taking advantage of the \gls{panoc} solving capabilities. Thus, a single cost function is built for the $M$ discrete scenarios, under a set of constraints. In such a manner, the solution of the multi-stage \gls{nmpc} problem satisfies all the scenarios and constraints. 

The simulation trials were running on a single core to solve the optimization problem and they performed in a personal computer with an Intel(R) Core(TM) i7-8550U @ $\unit[1.8]{Ghz}$ processor with $\unit[16]{GB}$ of RAM.


%
\section{Results} \label{sec:result}

\subsection{Simulation Setup} 
Out of the numerous simulations, three representative cases are selected to prove the performance of the proposed control scheme. For consistent results, data of delays generated from a \textit{Gamma} distribution $\Gamma(\alpha,\beta)$ with shape factor $\alpha=12$ and scale factor $\beta =0.015$ is used for all the simulation trials. During the simulations, the eight states of the \gls{mav} considered measurable and were updated based on the communication delays i.e. for the duration of a delay, the cloud computing controller has no knowledge of the states' evolution and the \gls{mav} maintains the last control input, while in \gls{mav}'s side (Fig.~\ref{fig:controlartichecture}), no delay is considered between low-level controller and the platform.

Two environments and three simulation trials are chosen to demonstrate the main attributes of the proposed control framework. The first environment includes a single large cylinder obstacle, where the performance of the multi-stage \gls{nmpc} is compared with the standard \gls{nmpc} under the presence of communication delays for navigating to a desired location, while the path is obstructed by the cylinder. By standard \gls{nmpc} is defined the solution of~\eqref{eq:optProb} for $M=1$, $\omega_i = 1$ and solved at nominal $t_s$. It should be noted that only the desired point is provided for the multi-stage \gls{nmpc} and \gls{nmpc} and the controller is generating collision free paths. For the second and the third simulation trials, a more complex environment is selected with three cylinders of different sizes. Same as in the first case, we compare the performance of the multi-stage \gls{nmpc} and the \gls{nmpc} under the presence of varying communication delays, while trying to reach a desired location avoiding collision with the three cylinders. Finally, we compare the performance of the multi-stage \gls{nmpc} and the \gls{nmpc} in the same environment without communication delays to demonstrate the better path planning of the proposed controller with a collision avoidance capability due to the consideration of higher sampling times.

\begin{figure}[htbp!] 
\setlength\fwidth{0.8\linewidth}
\centering
\includegraphics[width=0.8\linewidth]{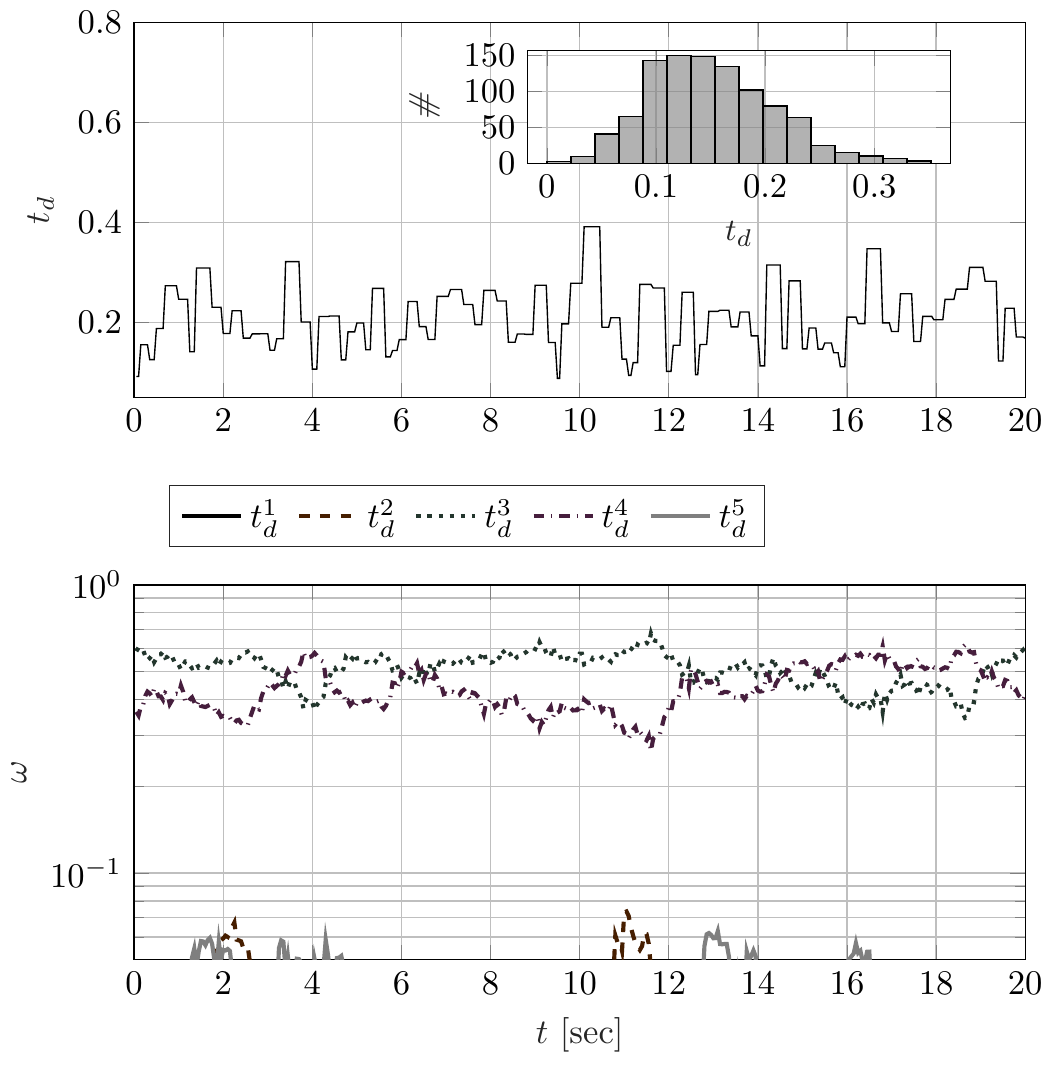}
\caption{Communication delays of the single cylinder obstacle simulation trial and histogram of the complete data of delays (Top). Calculated weights $\omega_i$ of the same simulation trial (Bottom).}
\label{fig:delay_data}
\end{figure}

Fig.~\ref{fig:delay_data} shows the delay data during the single obstacle simulation trial (Top), while inside the same illustration the histogram of the complete delays data-set is presented. Furthermore, the calculated probabilities for specific delays in logarithmic scale are given in Fig.~\ref{fig:delay_data} (Bottom). 

The parameters of the non-linear \gls{mav} model are the mass $m = \unit[1]{kg}$, the gravitational acceleration $g = \unit[9.81]{m/s^2}$, the mass normalized drag coefficients $A_x,A_y=0.1$ and $A_z =0.2$, and the time constants $t_{\phi},t_{\theta} = \unit[0.5]{sec}$ with gains $K_{\phi},K_{\theta} = 1$. The prediction horizon $N$ is 40, the nominal sampling time is $t_s = \unit[0.05]{sec}$, while the multi-stage \gls{nmpc} considers sampling times scenarios in seconds of $\bm{t_s} = [0.05, 0.07, 0.1, 0.2, 0.33]$. The roll and pitch angles are constrained within $[-\pi/18 \leq \phi, \theta \leq \pi/18],\ \unit{rad}$.
On account of the different simulation environments, the tuning weight parameters of the controller $\bm{Q}_x$ and $\bm{Q}_u$ vary among the simulations and the numerical values will be given in the sequel. 
\subsection{Single Cylinder Obstacle under Communication Delays}

\begin{figure}[htbp!] 
\setlength\fwidth{0.8\linewidth}
\centering
\includegraphics[width=0.8\linewidth]{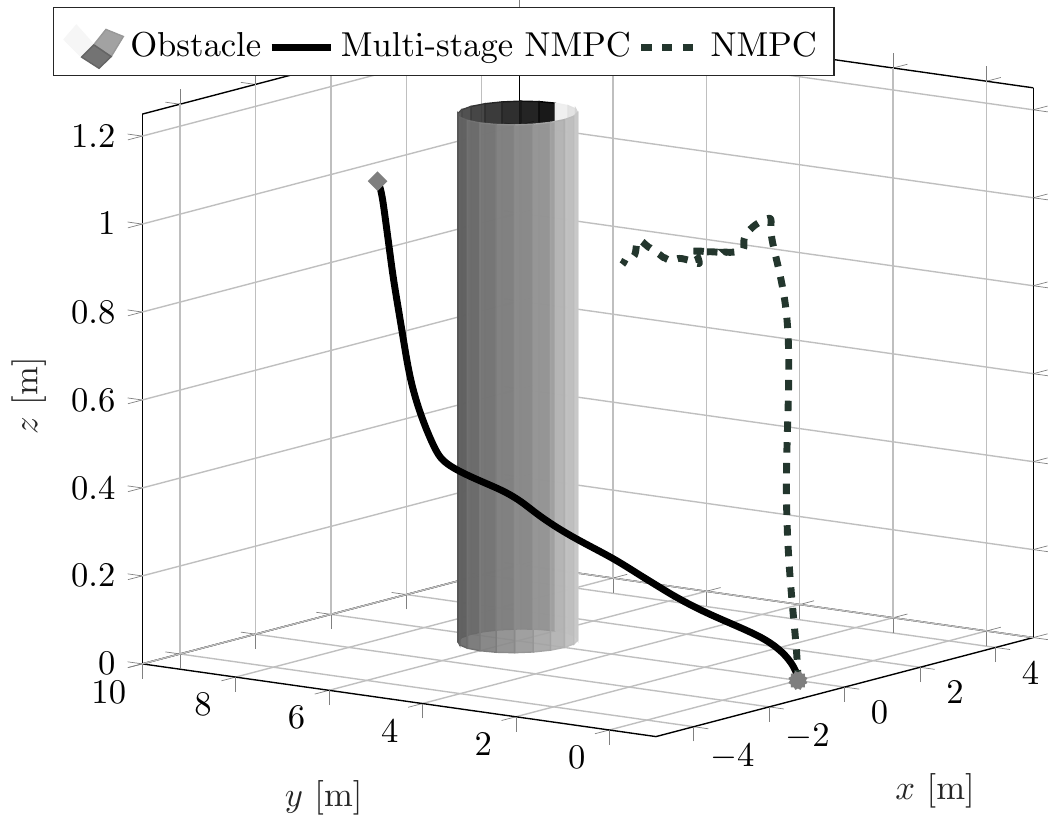}
\caption{Performance comparison of the multi-stage \gls{nmpc} versus \gls{nmpc} navigation under the effect of delays in presence of a cylinder obstacle}
\label{fig:3D1case}
\end{figure}

For the first case, as depicted in Fig.~\ref{fig:3D1case} a cylinder obstacle of radius $\unit[1.5]{m}$ and height $\unit[10]{m}$ located at $x,y~\unit[(0,6)]{m}$ is considered, while it is obstructing the straight path from the initial position $\bm{p}_{\text{init}}=[0,0,0],\ \unit{m}$ to the desire goal position $\bm{p}_{\text{goal}}=[0,9,1],\ \unit{m}$. For a better illustration, the obstacle in Fig.~\ref{fig:3D1case} is limited in the range of 0 to $\unit[1.25]{m}$. The weights of the states are $\bm{Q}_x = \text{diag}[6, 6, 20, 50, 50, 10, 20, 20]$, while the control action weights are $\bm{Q}_u = \text{diag}[20, 20, 20]$ and for the input rate of change are $\bm{Q}_{\Delta u} = \text{diag}[40, 65 ,65]$. 

In Fig~\ref{fig:3D1case} the multi-stage \gls{nmpc} successfully regulates the control inputs and the \gls{mav} navigates to the desired location despite the effect of delays. On the other hand, with the same weights and delays the \gls{nmpc} stops in a local minimal at approximately $\bm{p}=[0,5,1], \unit{m}$. Since, the multi-stage \gls{nmpc} considers higher sampling times, at most $t_s^5 = \unit[0.33]{sec}$ with the prediction horizon of 40 steps it can predict 13.2 seconds in the future when compared to the minimum sampling time scenario of $t_s^1 = 0.05$ seconds, which predicts at most up to the next 2 seconds.

\begin{figure}[htbp!] 
\setlength\fwidth{0.8\linewidth}
\centering
\includegraphics[width=0.8\linewidth]{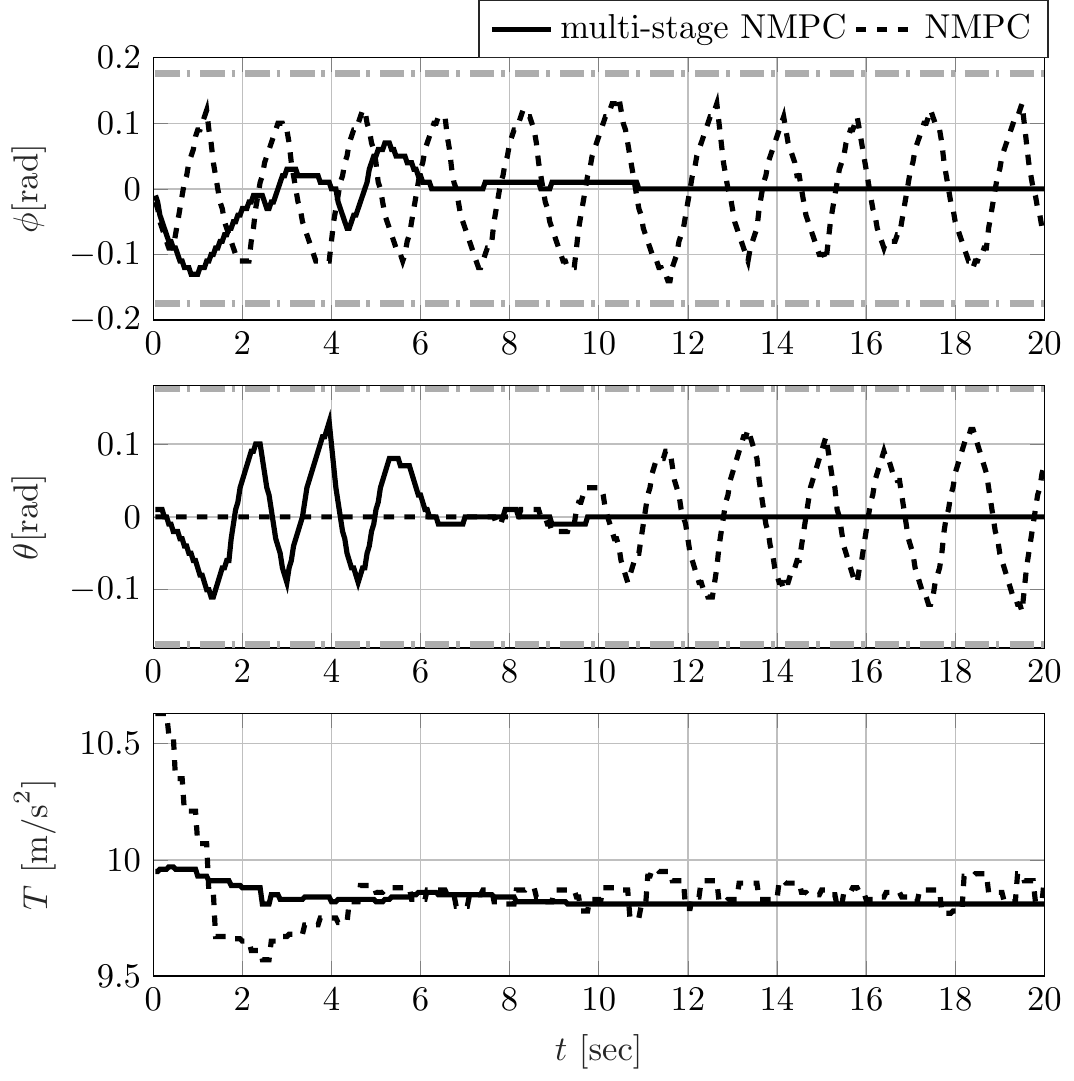}
\caption{Control action responses of roll, pitch and Thrust under the effect of communication delays in presence of a cylinder obstacle for multi-stage \gls{nmpc} and \gls{nmpc} }
\label{fig:RPT1case}
\end{figure}

Finally, for the first case the control actions roll pitch and thrust for both tested controllers is presented in Fig.~\ref{fig:RPT1case}. It can be noticed that the multi-stage \gls{nmpc} starts aggressively in the beginning but as approaches the desired location the roll and pitch angles are getting smoother. On the other hand, the \gls{nmpc} almost immediately falls into shacking issues due to the delays, something that could be observed in the Fig.~\ref{fig:3D1case} as well. Both the multi-stage  and classic \gls{nmpc} provide solutions that do not violate the given constraints denoted by light-grey dashed line in Fig.~\ref{fig:RPT1case}. For the thrust, the \gls{nmpc} is observed to be more aggressive when compared to the multi-stage \gls{nmpc} but both controllers successfully manage to regulate the height at $\unit[1]{m}$.

\subsection{Multiple Cylinder Obstacles under Communication Delays}

\begin{figure}[htbp!] 
\setlength\fwidth{0.8\linewidth}
\centering
\includegraphics[width=0.8\linewidth]{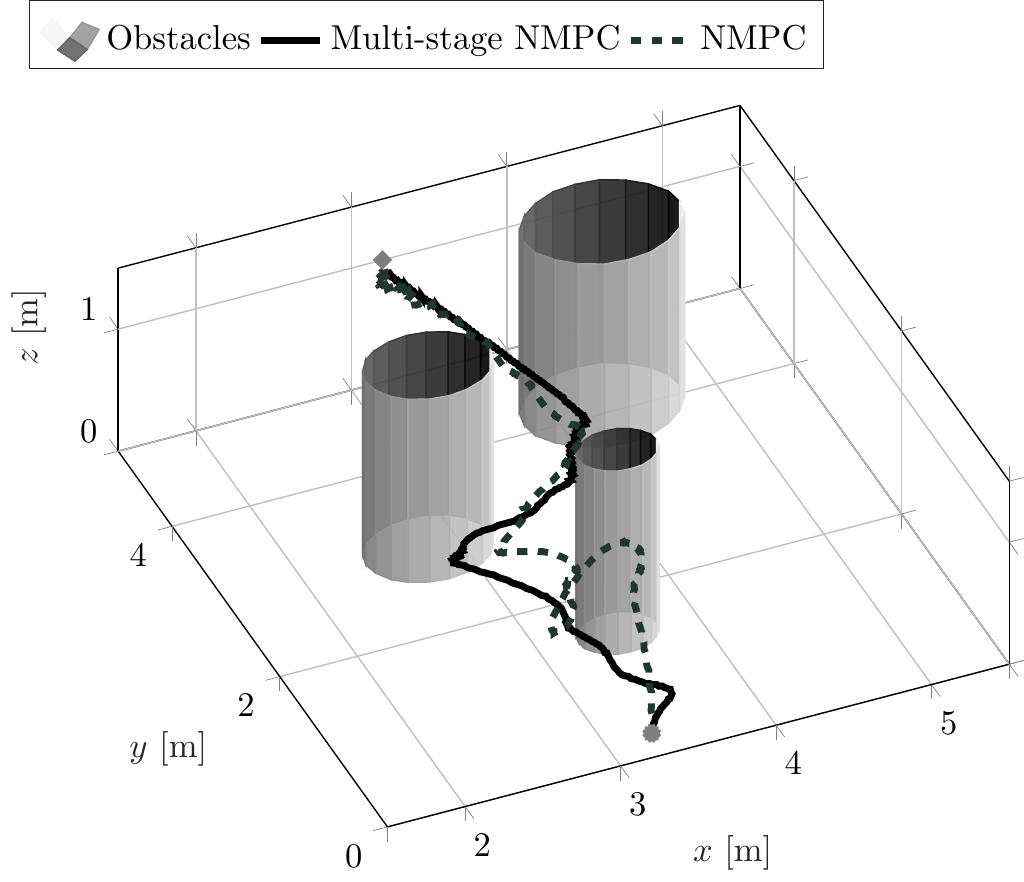}
\caption{Performance comparison of the multi-stage \gls{nmpc} versus \gls{nmpc} navigation under the effect of delays in presence of multiple cylinder obstacles}
\label{fig:3D_3cylinder_delaysON}
\end{figure}

For the second case three cylinder obstacles of radius 0.25, 0.4 and $\unit[0.5]{m}$ are located at $x,y~(3.5,1.5)~(2.8,3.0)~\text{and}~(4.3,4.1)\unit{m}$ respectively of $\unit[10]{m}$ height obstructing \gls{mav}'s path from the initial position $\bm{p}_{init}=[3.2,0,0]\,\unit{m}$ to the goal position $\bm{p}_{goal}=[3.2,5,1]\,\unit{m}$. As in the previous cases and for visualization purposes, the $z$-axis in Fig.~\ref{fig:3D_3cylinder_delaysON}, is limited between 0 and $\unit[1.25]{m}$. The tuning of the controllers is $\bm{Q}_x = \text{diag}[12.5, 12.5, 20, 30, 30, 10, 20, 20]$, while the control input weights are $\bm{Q}_u = \text{diag}[20, 20, 20]$ and the weight of the smoothness term is $\bm{Q}_{\Delta u} = \text{diag}[40, 165 ,165]$.

\begin{figure}[htbp!] 
\setlength\fwidth{0.8\linewidth}
\centering
\includegraphics[width=0.8\linewidth]{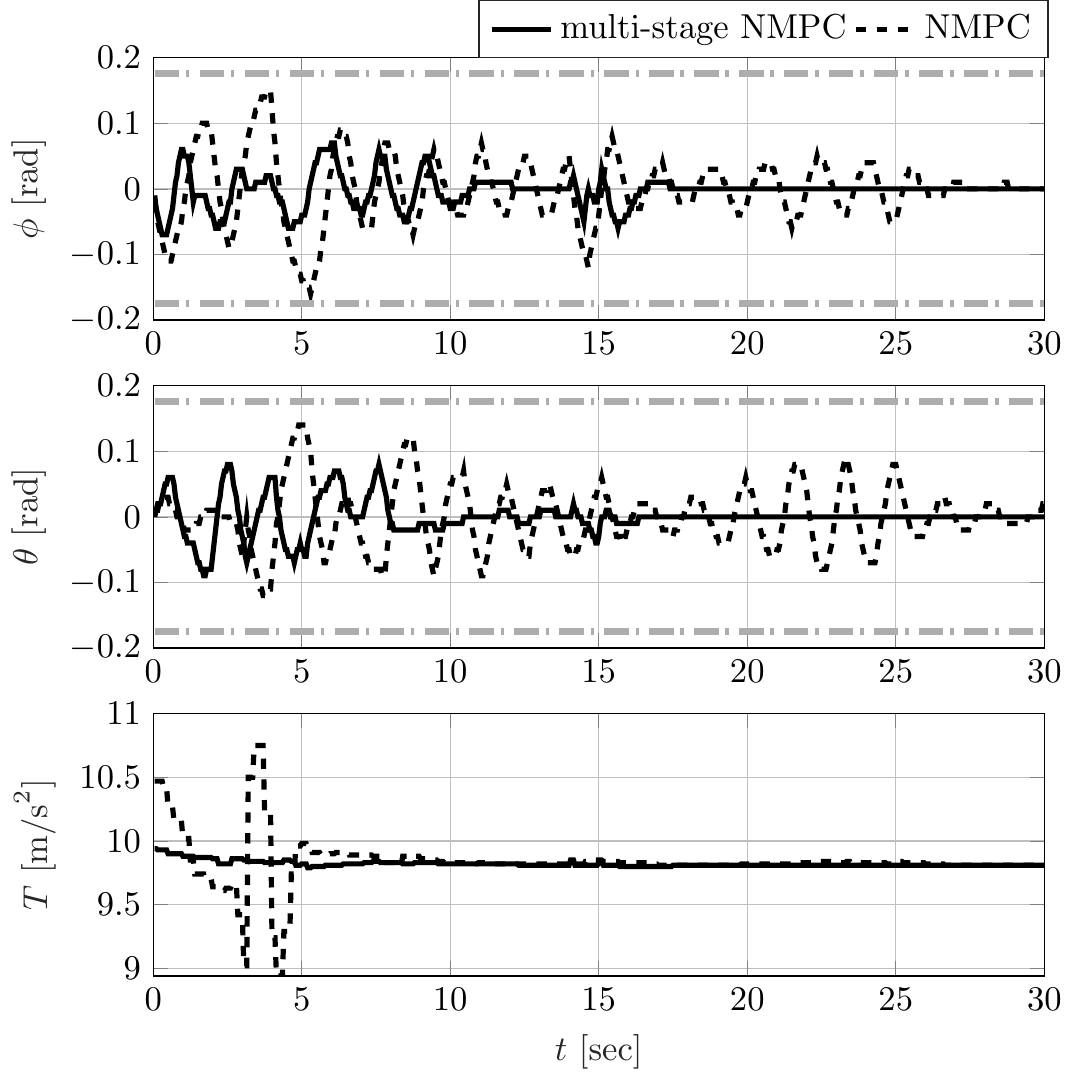}
\caption{Multi-stage \gls{nmpc} and \gls{nmpc} control actions roll, pitch and thrust under the effect of delays in presence of multiple cylinder obstacles.}
\label{fig:RPT_3cylinder_delaysON}
\end{figure}

Fig.~\ref{fig:3D_3cylinder_delaysON} illustrates the path followed by the multi-stage \gls{nmpc} (solid line) and the \gls{nmpc} (dashed line). Both controllers manage to navigate to the goal location. The multi-stage \gls{nmpc} path is characterized to be more smooth compared to the \gls{nmpc} path, something that can be observed by the smaller changes in $x,y$ and $z$ positions, as well as in the control actions in Fig~\ref{fig:RPT_3cylinder_delaysON}. The multi-stage controller manages to regulate the height steadily to the reference point in contrast to the \gls{nmpc}, which overshoots and undershoots in the beginning mainly due to the existence of the delays.

The roll, pitch and thrust commands of the second simulation are given in Fig.~\ref{fig:RPT_3cylinder_delaysON}. For the \gls{nmpc} the $\phi$ and $\theta$ are oscillating in the range $\pm\unit[0.1]{rad}$ for almost the entire time response. In contrast, the multi-stage \gls{nmpc} appears to be affected less by the communication delays and results in much smoother control actions. It is noticeable that \gls{nmpc} even if it manages to drive the \gls{mav} to the final position its path response is more aggressive as the \gls{mav} drifts from the expected position due to the delays. On the other hand, the multi-stage \gls{nmpc} results in a much smoother navigation path. This can be also identified in the altitude commands, where the control signal of the \gls{nmpc} changes abruptly at the beginning of the response and this is causing the platform to overshoot and undershoot in the height response as well.

\subsection{Multiple Cylinder Obstacles without Communication Delays}

\begin{figure}[htbp!] 
\setlength\fwidth{0.8\linewidth}
\centering
\includegraphics[width=0.8\linewidth]{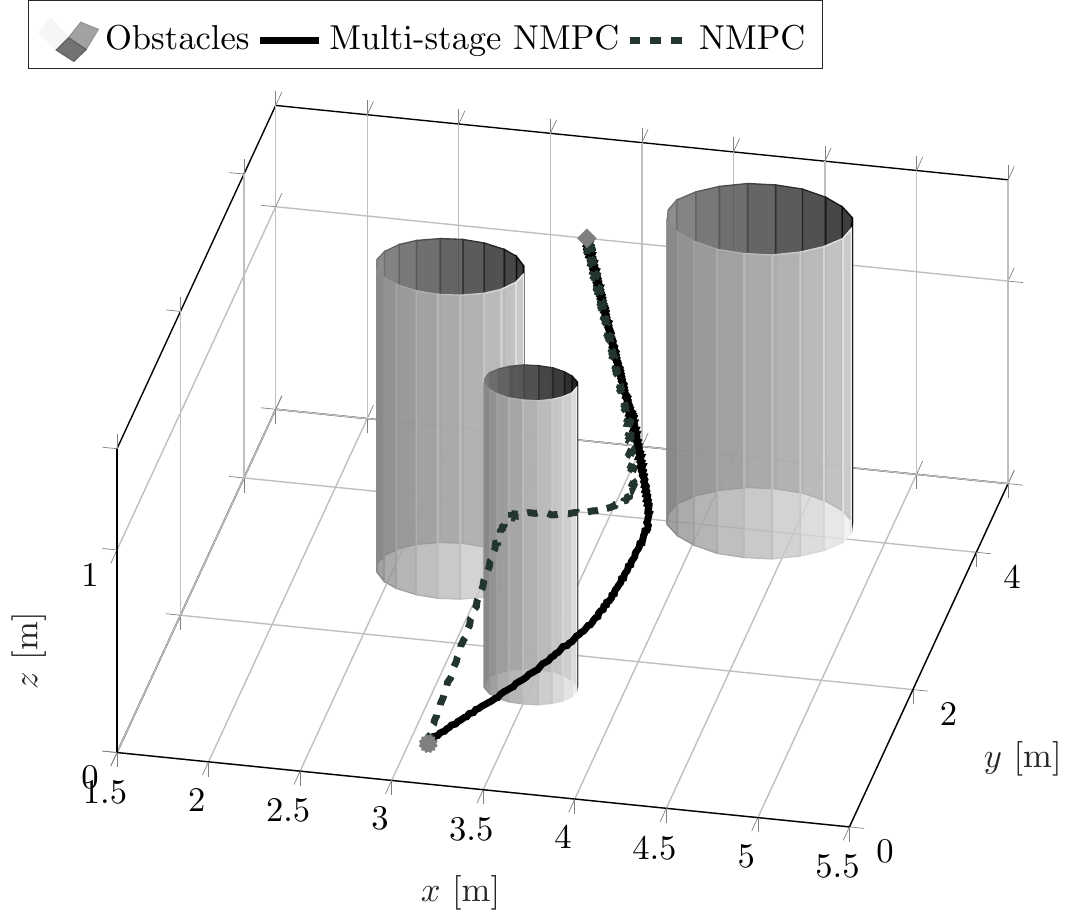}
\caption{Performance comparison of the multi-stage \gls{nmpc} versus the \gls{nmpc} navigation for $t_d =0$ in presence of multiple cylinder obstacles}
\label{fig:3D_3cylinder_delaysOFF}
\end{figure}

For the final test, the simulation tuning and constraints are identical to the second simulation, while no time delay is considered ($t_d=0$). The multi-stage probability is kept constant at $\bm{\omega}_i = [0.05, 0.15, 0.45, 0.30, 0.05]$ for the sampling times of $\bm{t}_s = [0.05, 0.07, 0.1, 0.2, 0.33], \unit{sec}$. Thus, the higher the sampling time is the further in the future we predict for achieving an improved path planning. As depicted in Fig.~\ref{fig:3D_3cylinder_delaysOFF}, under the absence of time delays both controllers successfully reach to the destination point with smooth maneuvers avoiding the obstacle from shorter path in comparison to the previous simulation. Even in this case, as it is presented in Table~\ref{tab:pathlentghcomparison}, the multi-stage \gls{nmpc} results in a shorter path compared to the \gls{nmpc}, while this time the difference between the two navigation performances is smaller.

\begin{figure}[htbp!] 
\setlength\fwidth{0.8\linewidth}
\centering
\includegraphics[width=0.8\linewidth]{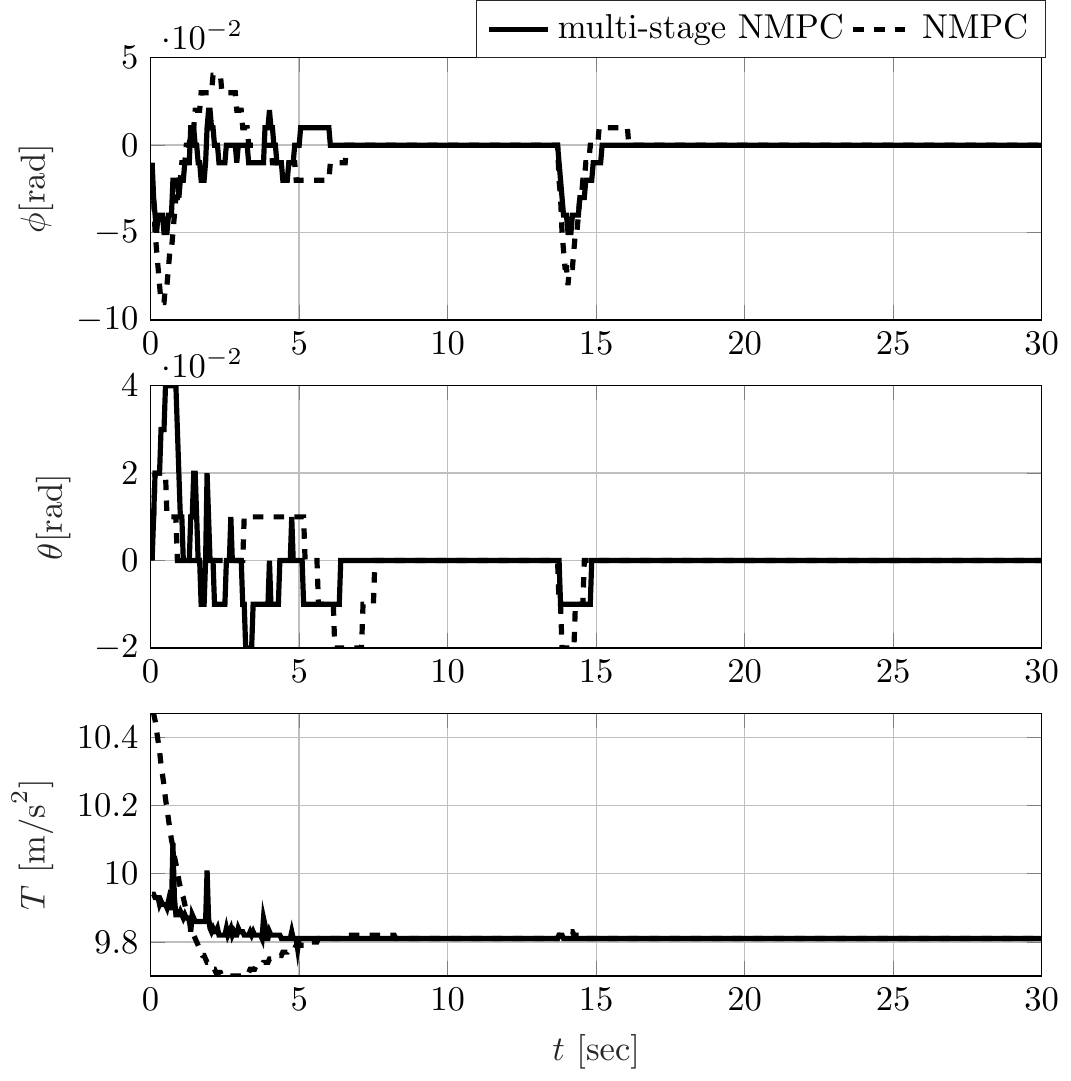}
\caption{Multi-stage \gls{nmpc} and \gls{nmpc} control actions of roll, pitch and thrust for $t_d=0$ in presence of multiple cylinder obstacles.}
\label{fig:RPT_3cylinder_delaysOFF}
\end{figure}

Fig.~\ref{fig:RPT_3cylinder_delaysOFF} presents the $\phi, \theta$ and $T$ actions of the last simulation under zero communication delays. The multi-stage \gls{nmpc} appears to be more resilient to the control input changes and thus resulting to smaller variations from the hover position. In comparison to Fig.~\ref{fig:RPT_3cylinder_delaysON} from the previous simulation, all the signal responses are smooth, indicating the intense effect of the  delays on the aerial platform.

Finally, in Tables~\ref{tab:pathlentghcomparison} and \ref{tab:solvertimecomparison} a comparison of the path length and mean computation time between the proposed multi-stage \gls{nmpc} is presented. For all the simulation trials, the distance from the initial point to the goal point is shorter for the multi-stage \gls{nmpc}. As far as the computation time is concerned, the \gls{nmpc} has a lower mean computation time as expected due to the smaller size of the optimization problem, but it must be highlighted that the \gls{nmpc} fails to overcome the obstacle in the first simulation and it has an overall lower performance in the second and third simulation. Furthermore, the computation time of the multi-stage \gls{nmpc} is fast enough since typical path planners operate at a sampling rate of $\unit[50]{ms}$. 

{\renewcommand{\arraystretch}{1.3}
\begin{table}[htbp!]
\caption{Comparison of the path length between the proposed multi-stage \gls{nmpc} and the path from \gls{nmpc}.}
\centering
\begin{threeparttable}[b]
\begin{tabular}{ccccc}
\hline
Scenario                & Simulation 1    & Simulation 2     & Simulation 3          \\ \hline
\gls{nmpc}              & 6.45\tnote{1} [m] &  8.99 [m]  & 5.75 [m]         \\ \hline
Multi-stage \gls{nmpc}  & 9.75 [m]    &  6.18 [m]  & 5.46 [m]         \\ \hline
\end{tabular}
     \begin{tablenotes}
     \item[1] Failed to reach final destination
   \end{tablenotes}
\end{threeparttable}
\label{tab:pathlentghcomparison}
\end{table}
}

{\renewcommand{\arraystretch}{1.3}
\begin{table}[htbp!]
\centering
\caption{Comparison of the average computation time between the proposed multi-stage \gls{nmpc} and the classic \gls{nmpc}.}
\begin{tabular}{ccccc}
\hline
Scenario                & Simulation 1    & Simulation 2     & Simulation 3          \\ \hline
\gls{nmpc}              & 15.1 [ms]    & 4.3 [ms]     & 1.6 [ms]          \\ \hline
Multi-stage \gls{nmpc}  & 20.5 [ms]    & 48.1 [ms]    & 19.7 [ms]          \\ \hline
\end{tabular}
\label{tab:solvertimecomparison}
\end{table}
}




\section{Conclusions} \label{sec:conclusions}
This article proposed a novel multi-stage \gls{nmpc} framework for collision-free navigation of \gls{mav} under the effect of time delays in the communication network.
The multi-stage \gls{nmpc} scenarios were based on different sampling times and varying weights derived from the communication delays. 
The proposed control scheme was evaluated under multiple simulations for different numbers of obstacles and variable communication delays. 
More specifically, the multi-stage was able to compensate for the network delays and navigate to the final point avoiding the cylinder obstacle that obstructed its path, while the classic \gls{nmpc} failed to reach the desired point.
The navigation in the environment with three obstacles without delays is considered, where the multi-stage \gls{nmpc} presented a smoother motion and fewer fluctuations in the control actions compared to the classic \gls{nmpc}, while with both controllers the \gls{mav} successfully navigated from the initial point to the final point.
For all the aforementioned cases, the generated paths were shorter compared to the fixed sampling rate \gls{nmpc}.
Lastly, even tough the mean computation time of the multi-stage controller was higher compared to the \gls{nmpc}, it is lower than shortest sampling time and the multi-stage control shows better performance in generating collision free paths.
Further studies will focus on the stability analysis and experimental evaluation with a 5G enabled \gls{mav} platform for collision-free navigation in urban environments.

\section{Declarations}
\subsection*{Funding}
This work has been partially funded by the European Union Horizon 2020 Research and Innovation Programme under the Grant Agreement No. 869379 illuMINEation. 
\subsection*{Conflict of interest/Competing interests}
The authors declare that they have no conflict of interest.
\subsection*{Availability of data and material}
Not applicable.
\subsection*{Code availability}
Not applicable.
\subsection*{Author's contributions}
\textbf{Andreas Papadimitriou:} Conceptualization, Methodology, Software, Validation, Formal analysis, Investigation, Writing - Original Draft, Visualization\\
\textbf{Hedyeh Jafari:} Conceptualization, Methodology, Validation, Resources, Writing - Original Draft\\
\textbf{Sina Sharif Mansouri: }Conceptualization, Methodology, Software, Validation, Resources, Writing - Original Draft, Visualization, Supervision, Project administration \\
\textbf{George Nikolakopoulos:} Funding acquisition, Supervision

\subsection*{Ethics approval}
Not applicable.
\subsection*{Consent to participate}
 Not applicable.
\subsection*{Consent for publication} 
Not applicable.

\bibliography{mybib.bib}

\end{document}

%% file: abbreviations.tex
\newacronym{ahmpc}{AHMPC}{Adaptive Horizon Model Predictive Control}
\newacronym{admm}{ADMM}{Alternating Direction Method of Multipliers}
\newacronym{ahe}{AHE}{Adaptive Histogram Equalization}
\newacronym{ar}{AR}{Augmented Reality}
\newacronym{cdf}{CDF}{Cumulative Distribution Function}
\newacronym{clahe}{CLAHE}{Contrast Limited Adaptive Histogram Equalization}
\newacronym{cnn}{CNN}{Convolutional Neural Network}
\newacronym{dbscan}{DBSCAN}{Density-Based Spatial Clustering of Applications with Noise }
\newacronym{dl}{DL}{Deep Learning}
\newacronym{dof}{DoF}{Degree of Freedom}
\newacronym{ekf}{EKF}{Extended Kalman Filter}
\newacronym{fbe}{FBE}{Forward Backward Envelope}
\newacronym{fov}{FoV}{Field of View}
\newacronym{fps}{fps}{Frame Per Second}
\newacronym{gnss}{GNSS}{Global Navigation Satellite System}
\newacronym{gps}{GPS}{Global Positioning System}
\newacronym{iid}{IID}{Independent Identically Distributed}
\newacronym{imu}{IMU}{Inertial Measurement Unit}
\newacronym{kf}{KF}{Extended Kalman Filter}
\newacronym{llwcf}{LLWCF}{Left Local Wall Continuum Function}
\newacronym{lrcn}{LRCN}{ Long-Term Recurrent Convolutional Network}
\newacronym{lwcf}{LWCF}{Local Wall Continuum Function}
\newacronym{mae}{MAE}{mean absolute error}
\newacronym{mav}{MAV}{Micro Aerial Vehicle}
\newacronym{mhe}{MHE}{Moving Horizon Estimation}
\newacronym{mimo}{MIMO}{Multiple Input Multiple Output}
\newacronym{mlp}{MLP}{MultiLayer Perceptron}
\newacronym{mpc}{MPC}{Model Predictive Control}
\newacronym{msf}{MSF}{Multi Sensor Fusion}
\newacronym{nmhe}{NMHE}{Nonlinear Moving Horizon Estimation}
\newacronym{nmpc}{NMPC}{Nonlinear Model Predictive Control}
\newacronym{nn}{NN}{Nueral Network}
\newacronym{nwu}{NWU}{North-West-Up}
\newacronym{open}{OpEn}{Optimization Engine}
\newacronym{panoc}{PANOC}{Proximal Averaged Newton-type method for Optimal Control}
\newacronym{pdf}{PDF}{Probability Density Function}
\newacronym{pid}{PID}{Proportional Integral Derivative}
\newacronym{psd}{PSD}{ Positive Semi-Definite }
\newacronym{rbf}{RBF}{ Radial Basis Function}
\newacronym{relu}{ReLU}{Rectified Linear Unit}
\newacronym{rl}{RL}{Reinforcement Learning}
\newacronym{rlwcf}{RLWCF}{Right Local Wall Continuum Function}
\newacronym{rmse}{RMSE}{Root Mean Square Error}
\newacronym{ros}{ROS}{Robot Operating System}
\newacronym{sfm}{SfM}{Structure from Motion}
\newacronym{spams}{SPAMS}{SPArse Modeling Software}
\newacronym{sqp}{SQP}{Sequential Quadratic Programming}
\newacronym{slam}{SLAM}{Simultaneous Localization and Mapping}
\newacronym{ugv}{UGV}{Unmanned Ground Vehicle}
\newacronym{ukf}{UKF}{Unscented Kalman Filter}
\newacronym{vi}{VI}{Visual Inertia}
\newacronym{vo}{VO}{Visual Odometry}